# CAN CONVERSATIONAL XAI IMPROVE USER PERFORMANCE? AN EXPERIMENTAL STUDY

*Short Paper*

Sven Kruschel, TU Dresden, Dresden, Germany, sven.kruschel@tu-dresden.de

Julian Rosenberger, University of Regensburg, Regensburg, Germany, julian.rosenberger@ur.de

Lasse Bohlen, TU Dresden, Dresden, Germany, lasse.bohlen@tu-dresden.de

Mathias Kraus, University of Regensburg, Regensburg, Germany, mathias.kraus@ur.de

Patrick Zschech, TU Dresden, Dresden, Germany, patrick.zschech@tu-dresden.de

## Abstract

*Explainable AI (XAI) techniques aim to provide insights into predictive models and enhance user performance, yet they often fall short of these expectations. Conversational XAI assistants promise to overcome such limitations, but empirical evidence on their impact on objective performance measures remains limited. We propose an experimental design for evaluating explanation assistance through prediction accuracy, model understanding, and error identification. Using an explainable-by-design prediction model, we create conditions where users can outperform the model by identifying and compensating for systematic errors. We compare conversational assistance against Q&A-based assistance to assess which better supports users in working with model explanations. Preliminary results from testing our experimental design show that participants (N=42) in both treatments significantly outperformed the model but reveal no performance differences between assistance types and modest engagement overall. These findings inform refinements for our planned full study, including enhanced engagement interventions and investigation of the mechanisms driving improved predictions.*

*Keywords: Explainable AI, Conversational XAI, Human-AI Collaboration, Error Identification.*

## 1 Introduction

As AI systems become more integrated into human decision-making, it is important to ensure that users can understand and rely on their predictions appropriately (Rudin, 2019). While many AI models achieve high overall performance, their predictions can still be unreliable in certain contexts, particularly when trained on limited or biased data. In such cases, human expertise can play a crucial role in complementing and improving algorithmic predictions. This recognition has driven the growing adoption of Explainable AI (XAI) techniques such as SHAP (Lundberg & Lee, 2017), LIME (Ribeiro et al., 2016), and explainable-by-design models like the Explainable Boosting Machine (EBM) (Lou et al., 2013), which provide transparency into model behavior through feature importance scores and feature contribution plots.

However, a critical gap exists between the promise and practice of XAI. While numerous techniques for generating explanations have been developed, there is limited empirical evidence on whether these explanations actually enable users to identify model errors and improve their decision-making performance (see e.g., Haag, 2026; Kaur et al., 2020; Suh et al., 2025). Most XAI evaluations focus on subjective metrics such as user satisfaction or perceived understanding, or assess performance with already-accurate models where error correction is not required. What remains underexplored specifically for conversational XAI is whether it helps users detect and compensate for systematic model failures (Buçinca et al., 2021).

Recent research has proposed conversational XAI interfaces, often powered by Large Language Models (LLMs), as a promising solution to enhance user engagement with model explanations. These systems enable users to engage with prediction models through natural language dialogue, ask follow-up questions, and explore model behavior interactively (Shen et al., 2023; Slack et al., 2023). Early evidence suggests conversational assistance improves user satisfaction and reduces time to find information compared to static dashboards (He et al., 2025). Yet, no prior work has directly examined whether LLM-based conversational explanation assistance improves users' ability to identify model errors and enhance their task performance when interacting with imperfect prediction models.

To address this gap, we propose an experimental study design for assessing how explanation assistance influences user performance through objective measures. Using a model trained on limited data, we create conditions where users can outperform the prediction model by identifying and compensating for systematic errors, enabling us to test whether conversational assistance helps users understand model limitations and improve their predictions compared to static Questions and Answers (Q&A)-based support.

We collected initial data through an online study (N = 42) in which participants worked with an explainable-by-design model for bike rental predictions trained on limited data. Both groups had access to the model's predictions and plots showing how each feature contributes to predictions.

The *Q&A* treatment provided structured written guidance for interpreting these visualizations, while the *conversational* treatment offered an interactive assistant that could answer questions in natural language. We evaluated performance through three objective measures: *prediction accuracy*, *model understanding*, and *error identification*.

Preliminary results validate the experimental design while revealing key considerations for our planned full study. Participants in both treatments significantly outperformed the model, but we observed no performance differences between assistance types and modest assistance engagement overall. These patterns inform refinements for our full study, including enhanced engagement interventions and investigation of the mechanisms driving improved predictions. We present this experimental design and initial findings to establish a foundation for discussion on how to better isolate when conversational assistance adds value.

The remainder of this paper is structured as follows. Section 2 reviews related work on conversational XAI. Section 3 describes our proposed experimental study design and the objective measures. Section 4 presents preliminary findings. Section 5 discusses design modifications for our planned full study.

# 2 Foundations and Related Work

XAI encompasses a range of explanation types, including counterfactual explanations and example-based approaches. In prediction tasks like ours, interfaces commonly rely on feature-importance techniques such as SHAP, LIME, and explainable-by-design models like the EBM to generate explanations highlighting which features influenced predictions (Kruschel et al., 2026). These explanations are typically presented through static dashboards. Working with such dashboards in prediction tasks is itself a search activity: users may engage in focused search to answer specific questions about a prediction, or scan across explanation views to identify useful patterns and challenge assumptions (Vandenbosch & Huff, 1997). In our setting, this means inspecting feature contribution plots and model predictions to identify regions where the model may behave unexpectedly (Kaur et al., 2020). Yet static dashboards often fail to support the dynamic, iterative questioning required for effective human-AI collaboration. Users cannot easily ask comprehension and follow-up questions, explore alternative scenarios, or probe the model's behavior in specific regions of interest, limitations that hinder their ability to identify when the model might fail.

Recent research explores conversational XAI as a more interactive alternative. These systems enable users to query prediction models and request explanations through natural language dialogue, fundamentally shifting from passive visualization consumption to active explanation exploration (e.g., Bordt et al., 2024; Slack et al., 2023), which aligns with the idea that effective decision support should let users probe model behavior and test their own hypotheses about errors (Miller, 2023). Initial conversational XAI evidence suggests faster model-related question answering as well as higher objective understanding and trust than static interfaces (Slack et al., 2023; Mindlin et al., 2024; He et al., 2025). Explanations may also support learning from AI feedback, so conversational follow-up could aid learning more than static formats (Förster et al., 2025).

However, these improvements in usability and perceived understanding come with substantial risks. XAI interfaces, both traditional dashboards and conversational assistance, have been shown to promote overreliance, where users accept algorithmic recommendations too readily without sufficiently critical evaluation (Buçinca et al., 2021; He et al., 2025). This overreliance is often attributed to the illusion of explanatory depth, a cognitive bias where the presence of explanations leads users to overestimate their actual understanding of how the prediction model works (Chromik et al., 2021), a risk compounded by evidence that explanations can reshape users' information processing in ways subject to confirmation bias, allowing misconceptions to persist (Bauer et al., 2023). Conversational XAI assistants may exacerbate this risk further, as LLMs' linguistic sophistication could strengthen the illusion while undermining users' objective understanding (He et al., 2025).

In summary, while conversational XAI demonstrates advantages in usability and engagement, it introduces new risks of miscalibrated trust and illusory understanding. Achieving appropriate reliance, a state where users trust algorithmic recommendations when warranted but maintain critical judgment when necessary (Buçinca et al., 2021), requires careful design choices that prioritize transparent communication of model limitations over mere fluency or persuasiveness (He et al., 2025).

In our task, users must search the dashboard, turn uncertainty into questions, and probe model behavior. We therefore expect conversational assistance to outperform fixed Q&A as it can adapt to case-specific information needs, reducing search effort, and supporting learning from feedback (Förster et al., 2025; Miller, 2023; Vandenbosch & Huff, 1997). We thus expect higher model understanding, error identification, and prediction accuracy compared to fixed Q&A support. This way, our study contributes to this literature by examining whether conversational XAI assistance improves objective performance measures in conditions where users must identify and compensate for model errors.

# 3 Methods

This section describes our experimental design for evaluating explanation assistance through objective performance measures. Participants work with an explainable-by-design model for bike rental predictions trained on a limited subset of data. Depending on the treatment, they receive either LLM-based conversational or Q&A-based assistance. We evaluate user performance through prediction accuracy, model understanding, and error identification.

The following subsections detail the task and model setup, treatments, procedure, and measures. For more details about how the study was implemented please refer to the online supplementary materials.[1]

## 3.1 Task and Model Setup

To examine how conversational XAI assistance influences user performance, we use a prediction task based on the UCI Bike Sharing Dataset.[2] This dataset includes hourly bike rental counts together with weather- and calendar-related attributes. In the experiment, participants are asked to predict the number of bike rentals on a 0 – 1000 scale given these features.

The predictive model used in this task is an Explainable Boosting Machine (EBM) (Lou et al., 2013), an explainable-by-design model that provides feature-level visualizations illustrating how each feature contributes to the final prediction. We choose this model because it offers a strong balance between predictive performance and full explainability (Kruschel et al., 2026; Rudin, 2019).

[1] https://osf.io/84ft7/overview

[2] https://archive.ics.uci.edu/ml/datasets/Bike+Sharing+Dataset

To make model errors observable, we randomly undersample the training data. This raises MAE from ≈ 80 to ≈ 130 and more obvious implausible patterns emerge, e.g., higher predicted demand at high wind speeds. Rather than simulating a deployment-ready model, this setup provides an experimental environment in which systematic prediction errors occur frequently enough to be observed by participants within the time constraints of an online survey.

With the bike rental scenario, we intentionally chose a domain with interpretable and commonly understood features like temperature, wind speed, humidity, time of day, type of day, and season. This allows participants without domain expertise to reason about the predictions and potentially detect model errors. This design enables us to study how lay decision makers interact with and critically evaluate model predictions that conflict with intuitive expectations while keeping the task manageable within the time constraints of an online experiment.

By comparing predictions from the undersampled model with those from a model trained on the full dataset, we can further examine whether participants are able to identify feature regions in which the lower-quality model deviates most strongly. This setup allows us to study whether conversational explanation assistance helps users recognize, interpret, and account for model errors when making their own predictions.

## 3.2 Experimental Design and Treatments

The online experiment is conducted via Prolific, and 42 participants are recruited, filtered for at least one year of work experience and professional roles involving decision-making in marketing, operations, or related industries (e.g., manufacturing, healthcare, or consulting). We consider these qualifications to be the minimum requirements that would be expected in a real-world scenario for interacting with such a model in this type of prediction task. All participants received a base payment of £7.46. This was for the 74 minutes they spent on average. To incentivize serious engagement, we also paid a performance-based bonus of up to £8.00. To examine the influence of different explanation assistance types, participants are randomly assigned to one of the two treatments: a conversational treatment and a Q&A treatment.

In the conversational treatment, participants have the possibility to interact with an LLM-based conversational assistant designed to help them interpret the model's explanations. To implement this setup, we developed a conversational agent based on a pretrained OpenAI LLM, equipped with functions that allow it to access the underlying prediction model and respond to users' prediction-related questions (see supplementary materials for implementation details).

In the Q&A treatment, participants have access to a static Q&A that provides structured written guidance on how to interpret the model explanations. The Q&A entries were derived from common pretest questions and topics the conversational assistant was designed to cover. Because all participants used the same EBM dashboard, the manipulation concerns rather the form of additional assistance: fixed guidance versus interactive, context-specific support than the presence of explanations themselves.

The study consists of multiple sequential phases in which participants must interact with the prediction model, explore its explanations, and can request assistance according to their assigned treatment. In the following, we describe each phase in order (see Figure 1 for an overview of the experimental interface).

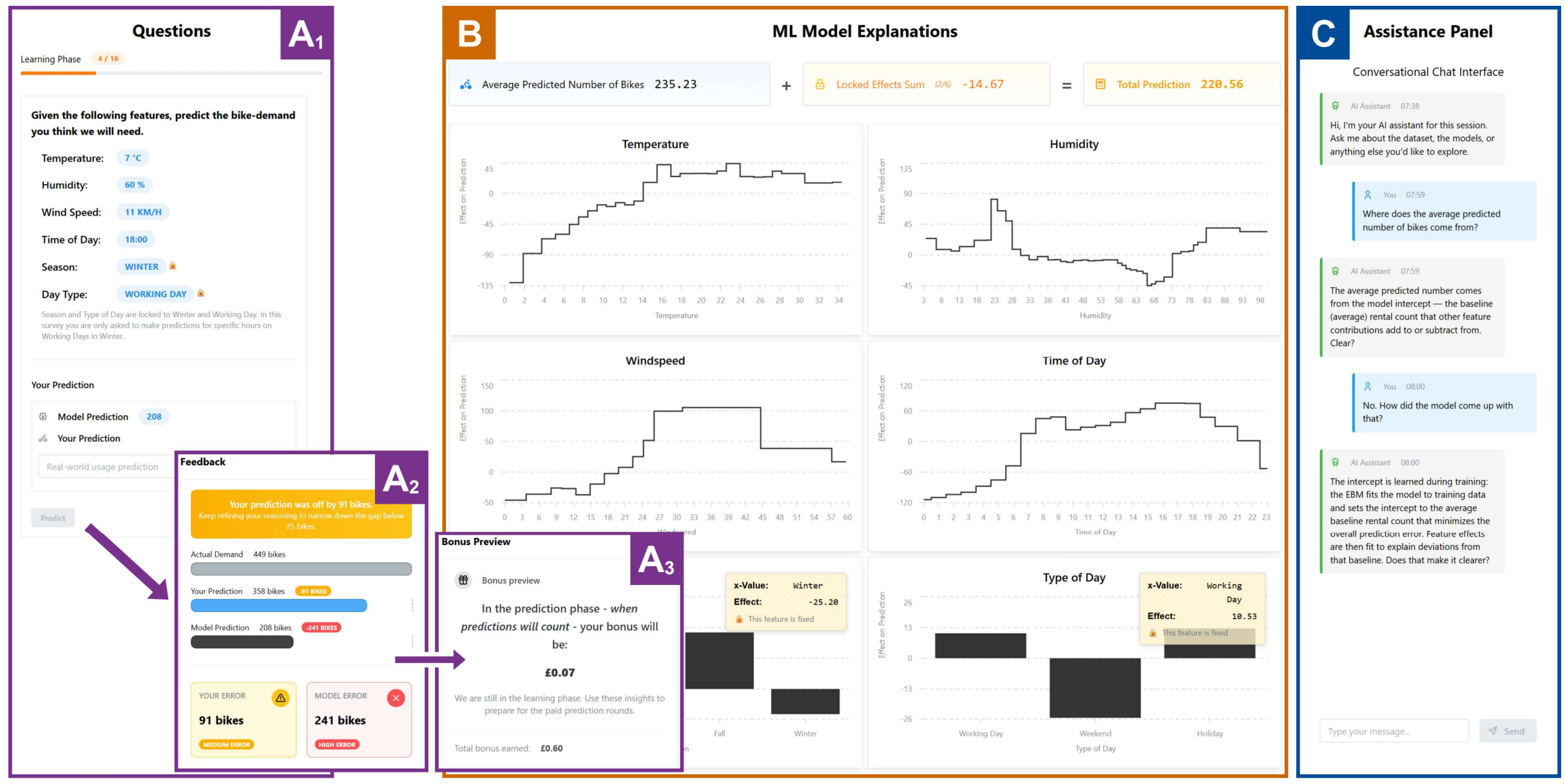

*Figure 1. Main elements that participants worked with. ($A_1$) The question sidebar with questions, feature values, and prediction input. ($A_2$) Feedback display after participants submitted their predictions. ($A_3$) Bonus display showing the potential bonus. (B) Model explanation section with contribution plots showing each feature's effect on predicted bike rentals. (C) Assistant panel to retrieve additional information or guidance.*

*Introduction Phase.* Participants are first introduced to the purpose of the study, the prediction task, and the functionality of the EBM prediction model. They then complete a brief tutorial using example feature contribution plots to familiarize themselves with how the explanations work. Afterwards, they are informed that the model used in the main task is imperfect, as reflected in its MAE, and that the bike-sharing service has experienced issues with inaccurate predictions. Finally, participants are informed about the performance-based bonus and the type of assistance available according to their assigned treatment.

*Learning Phase.* In the subsequent learning phase, participants make predictions and explore the model's explanations by examining graphs that display each feature's contribution to the prediction (Figure 1 B). This process allows them to develop an understanding of how the model works. By predicting bike demand and receiving feedback, participants can identify cases where the model's predictions deviated from reality. During this phase, they view both the model's predictions and the corresponding feature contribution plots, and then enter their own predictions (Figure 1 $A_1$). Sixteen samples are selected for this phase – eight that the model predicts accurately and eight where it performs poorly. The poorly predicted samples are chosen because their feature values fell in regions where the undersampled model diverged substantially from a model trained on the full dataset. Participants also receive feedback on the model's predictions, their own prediction accuracy, and the potential bonus they can earn if they achieved similar accuracy in the subsequent prediction phase (Figure 1 $A_2$ & $A_3$). During the interaction, depending on their assigned treatment, participants could request additional information either through a Q&A panel or via LLM-based assistance (Figure 1 C). This assistance was designed to help participants interpret the model's explanations more effectively and to encourage deeper reflection on the model's reasoning throughout the learning process.

*Prediction Phase.* In the subsequent prediction phase, participants make predictions for a new but comparable set of samples in randomized order. Unlike in the learning phase, they do not receive immediate feedback on their predictions (feedback panels $A_2$ and $A_3$), nor do they have access to the explanation assistance (Figure 1 C). Furthermore, whereas in the learning phase participants are only informed about the bonus they would have earned based on their prediction accuracy, in this phase they actually receive a performance-based bonus. This design encourages careful application of the model insights gained earlier, as prediction accuracy now directly determines their financial reward. At the end of the prediction phase, participants complete two sets of eight single-choice questions to assess their model understanding and error identification skills. As assistance is unavailable at this phase, these measures capture actual retained learning rather than their pure ability to query the assistance.

*Post-Survey.* After completing the prediction phase, participants fill out a post-task survey. The survey includes multiple constructs drawn from prior conversational XAI literature (e.g., trust, satisfaction, cognitive load, AI literacy) to capture participants' perceptions and experiences (see supplementary materials). Participants also provide feedback on their interaction with the system, reflecting on how they experience the explanations and the assistance used in their treatment.

## 3.3 Measures

Our study assesses the effect of explanation assistance type across three key dimensions: prediction accuracy, model understanding, and error identification. These measures jointly capture how participants perform during the task, understand the model, and recognize areas of model weaknesses. Because prediction accuracy may reflect both domain intuition and learned model logic, we complement it with two comprehension-based measures.

*Prediction Accuracy.* This dimension reflects the overall outcome of participants' judgments when working with the model. It is measured by the MAE of participants' final predictions compared to the true bike rental counts. Achieving a higher predictive performance than the undersampled model requires participants to recognize and account for its systematic errors when making their own predictions.

*Model Understanding.* This dimension reflects participants' comprehension of the model's structure and underlying logic. It is assessed using single-choice questions (1 out of 5 options) that test their understanding of the model's functionality and the relationships shown in the feature contribution plots. The correct answers are determined by the model's actual contribution patterns, meaning that participants must identify how the model behaves in specific feature regions rather than rely only on general intuition about bike demand.

*Error Identification.* This dimension evaluates participants' ability to recognize where the model performs poorly. It is measured through single-choice questions (1 out of 6 options) in which participants indicate the feature interval they believe corresponds to a substantial model error. The correct intervals are defined as the regions where the undersampled model diverges most from a model trained on the full dataset. This operationalization focuses on systematic errors in specific feature regions rather than isolated observations. Because these regions are derived from the model's actual deviations rather than from general beliefs about the domain, the measure is intended to capture recognition of model-specific weaknesses. Accurately identifying these regions demonstrates an understanding of the model's limitations and serves as a prerequisite for effectively improving predictive performance.

# 4 Preliminary Results

This section reports preliminary findings focusing on the three objective dimensions of user performance. While additional constructs were measured, their analysis lies beyond the scope of this paper and will be reported in subsequent work.

The results reveal that participants in both treatments have a solid awareness for the task and were able to effectively adjust their predictions based on model insights. A one-sample t-test comparing participants' prediction errors to the undersampled model (MAE = 120.56) showed that participants achieved significantly lower errors (MAE = 99.31, $t(41) = -6.96$, $p < .001$). This confirms that participants outperformed the prediction model and were able to make meaningful corrections. Moreover, both groups performed clearly above random chance levels in the comprehension-based measures. With five answer options for the model understanding questions (chance level = 0.20) and six for the error detection questions (chance level = 0.17), average accuracies around 0.48 and 0.33 respectively indicate that participants meaningfully interpreted the explanatory visualizations and recognized regions of model weakness. Figure 2 shows the distributions of all three measures as box plots.

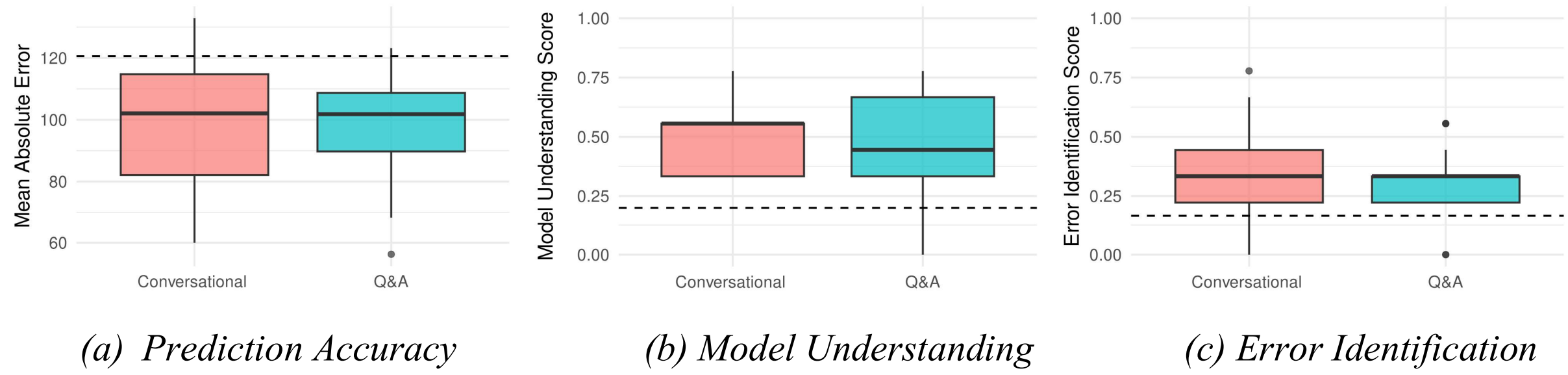

*(a) Prediction Accuracy* *(b) Model Understanding* *(c) Error Identification*

*Figure 2. Performance measures across treatments. Lower values in prediction accuracy represent smaller mean absolute error (MAE). Scores for model understanding and error identification indicate the share of correct answers. Dashed lines show baseline performance: undersampled model's MAE (120.56) for prediction accuracy and chance levels (0.20 and 0.17) for the two score measures, respectively.*

After establishing that participants generally engaged effectively with the task, we next examined potential differences between assistance treatments. Overall, only small and statistically insignificant differences were observed between both treatment groups across all three dimensions. For model understanding, a Welch two-sample t-test indicated no significant difference between the conversational (M = 0.50) and Q&A (M = 0.47) assistance groups ($t(36.93) = 0.49$, $p = .63$). For error identification, there was also no significant difference between groups ($t(38.21) = 0.60$, $p = .55$), with mean scores of 0.34 (conversational) and 0.31 (Q&A). Finally, for prediction accuracy, no significant group differences emerged ($t(36.88) = 0.46$, $p = .64$), with mean MAE values of 100.73 (conversational) and 97.87 (Q&A). Taken together, these findings suggest that, with our current experimental design, LLM-based conversational assistance does not yield measurable performance improvements compared to static information support. Given the small observed effect size, we would not expect this result to change substantially even with a much larger participant sample.

# 5 Discussion

This research in progress makes three key contributions to conversational XAI evaluation. We propose an experimental design that evaluates user performance through objective measures rather than subjective measures such as user satisfaction (He et al., 2025). To measure prediction accuracy, we simulate conditions in which human expertise is required to enhance model performance by using a model trained on limited data. We also introduce measures to evaluate participants' model understanding and their ability to identify areas of model weakness.

Further, this design can be replicated across domains where model limitations need to be identified and compensated for, enabling assessment of whether conversational XAI can support this process. Additionally, our comparison of conversational versus Q&A explanation assistance provides initial empirical evidence on whether LLM-based conversational assistance offers measurable advantages over static information in this context. Finally, our preliminary findings confirm that participants can outperform the undersampled model. Nevertheless, our preliminary findings reveal room for improvement for the design of the full study and raise important questions about conversational XAI effectiveness. We discuss these implications in terms of four key observations that emerged from our initial data and their implications for ongoing research.

First, participants achieved significantly better prediction accuracy than the undersampled model (MAE 100.73 vs. 120.56) yet showed only modest performance on model understanding (47–50%) and error identification (32–34%). This raises a critical question: did participants develop true conceptual understanding, or did they simply learn to apply task-specific rules that yielded better predictions without real comprehension? While our question-based measures were intended to capture the model's logic and model-specific weaknesses, the current design cannot fully disentangle whether participants learned the model, relied on domain intuition, or developed task-specific rules. Accordingly, prediction accuracy should be interpreted as an overall performance outcome rather than as direct evidence of true model comprehension. This distinction matters for evaluating XAI user performance and overreliance risks, as users may overestimate their understanding when presented with explanations (Chromik et al., 2021) and develop inappropriate trust (Buçinca et al., 2021). Relatedly, the equal split of accurately and poorly predicted samples, chosen to ensure sufficient error exposure within the study's time constraints, may have skewed participants' perceptions of model reliability and thereby influenced their reliance behavior. We will therefore modify feedback mechanisms to emphasize model understanding and error identification, not just prediction accuracy, refine the comprehension questions to better distinguish model logic from domain intuition, and adopt a more realistic error rate in the planned full study, promoting deeper engagement to determine whether performance gains reflect true comprehension or task-specific adaptations.

Second, our preliminary results reveal no significant differences between conversational and Q&A explanation assistance across all three performance measures, suggesting the conversational assistance provided no measurable advantages. Several mechanisms may explain this finding. Participants may have underutilized the conversational features – despite spending an average of 74 minutes on the experiment – due to limited engagement incentives, as users often default to passive consumption rather than active exploration (Kaur et al., 2020). Alternatively, the task itself may have been too straightforward to prompt exploratory questioning (Miller, 2019), or the transparent model structures of the EBM may have already provided sufficient explainability for the prediction task. To better identify the conditions under which conversational assistance adds value, we plan to shorten the initial model introduction phase and introduce a black-box treatment to disentangle the effects of explanations from those of feedback mechanisms.

Third, for the planned full study, we will systematically decompose study design elements by varying feature contribution plots, assistance type (none, Q&A, or conversational), and feedback mechanisms. This factorial approach will determine which components are necessary for improved decision-making, including a no-assistance baseline to establish whether explanation assistance improves performance beyond feedback alone. By analyzing interaction logs, we can examine whether deeper engagement with conversational assistance yields different learning trajectories. This ongoing research aims to clarify when LLM-based conversational assistants help users, improve model performance, and provide actionable guidance for XAI design.

Fourth, several contextual factors may moderate conversational XAI effectiveness (Miller, 2019). User expertise, domain familiarity, decision stakes and time pressure may influence whether participants benefit more from structured Q&A information or conversational assistance. Especially, task complexity is likely to play a crucial role, given that our simplified, controlled setup was designed for an online study and differs from real-world deployments. Real-world scenarios typically involve more complex models that require substantial time and domain expertise to analyze model behavior and detect errors. Future work will systematically examine these moderating factors.

Until then, the question of whether conversational assistance enhances objective user performance remains open, particularly when well-designed alternatives exist. Our ongoing work aims to establish whether conversational XAI assistance can deliver such improvements, and under what conditions such improvements occur.